\documentclass[11pt,a4paper]{article}
\usepackage[hyperref]{emnlp-ijcnlp-2019}
\usepackage{times}
\usepackage{latexsym}

\usepackage{url}
\usepackage{multirow}
\usepackage{amsmath}
\usepackage{booktabs}
\usepackage{bigstrut,bigdelim}
\usepackage{amssymb}
\usepackage{graphicx}
\usepackage{subfigure}
\usepackage{bm}
\usepackage{endnotes}

\aclfinalcopy 

\title{NCLS: Neural Cross-Lingual Summarization}

\author{
  Junnan Zhu$^{1,2}$,
  Qian Wang$^{1,2}$,
  Yining Wang$^{1,2}$,\\
  \textbf{Yu Zhou}$^{1,2\thanks{\ \ Corresponding author.}}$,
  \textbf{Jiajun Zhang}$^{1,2}$,
  \textbf{Shaonan Wang}$^{1,2}$,
  \and \textbf{Chengqing Zong}$^{1,2,3}$
  \\
  $^1$ National Laboratory of Pattern Recognition, Institute of Automation, CAS, Beijing, China\\
  $^2$ University of Chinese Academy of Sciences, Beijing, China\\
  $^3$ CAS Center for Excellence in Brain Science and Intelligence Technology, Beijing, China\\
  {\tt \{junnan.zhu, yzhou, jjzhang, cqzong\}@nlpr.ia.ac.cn}
}

\date{}

\begin{document}
\maketitle
\begin{abstract}
Cross-lingual summarization (CLS) is the task to produce a summary in one particular language for a source document in a different language. Existing methods simply divide this task into two steps: summarization and translation, leading to the problem of error propagation. To handle that, we present an end-to-end CLS framework, which we refer to as Neural Cross-Lingual Summarization (NCLS), for the first time. Moreover, we propose to further improve NCLS by incorporating two related tasks, monolingual summarization and machine translation, into the training process of CLS under multi-task learning. Due to the lack of supervised CLS data, we propose a round-trip translation strategy to acquire two high-quality large-scale CLS datasets based on existing monolingual summarization datasets. Experimental results have shown that our NCLS achieves remarkable improvement over traditional pipeline methods on both English-to-Chinese and Chinese-to-English CLS human-corrected test sets. In addition, NCLS with multi-task learning can further significantly improve the quality of generated summaries. We make our dataset and code publicly available here: \url{http://www.nlpr.ia.ac.cn/cip/dataset.htm}.
\end{abstract}

\section{Introduction}
Given a document in one source language, cross-lingual summarization aims to produce a summary in a different target language, which can help people efficiently acquire the gist of an article in a foreign language. Traditional approaches to CLS are based on the pipeline paradigm, which either first translates the original document into target language and then summarizes the translated document~\cite{leuski2003cross} or first summarizes the original document and then translates the summary into target language~\cite{lim2004multi, orasan2008evaluation, wan2010cross}. However, the current machine translation (MT) is not perfect, which results in the error propagation problem. Although end-to-end deep learning has made great progress in natural language processing, no one has yet applied it to CLS due to the lack of large-scale supervised dataset.

The input and output of CLS are in two different languages, which makes the data acquisition much more difficult than monolingual summarization (MS). To the best of our knowledge, no one has studied how to automatically build a high-quality large-scale CLS dataset. Therefore, in this work, we introduce a novel approach to directly address the lack of data. Specifically, we propose a simple yet effective round-trip translation strategy to obtain cross-lingual document-summary pairs from existing monolingual summarization datasets~\cite{hermann2015, Zhu2018MSMO, lcsts}. More details can be found in Section~\ref{dataconstruction} below.


Based on the dataset that we have constructed, we propose end-to-end models on cross-lingual summarization, which we refer to as Neural Cross-Lingual Summarization (NCLS). Furthermore, we consider improving CLS with two related tasks: MS and MT. We incorporate the training process of MS and MT into that of CLS under the multi-task learning framework~\cite{caruana1997multitask}. Experimental results demonstrate that NCLS achieves remarkable improvement over traditional pipeline paradigm. In addition, both MS and MT can significantly help to produce better summaries.
\begin{figure*}[t]
\setlength{\abovecaptionskip}{0.12cm}    
\setlength{\belowcaptionskip}{-0.3cm}
  \centering
  \includegraphics[width=1\linewidth]{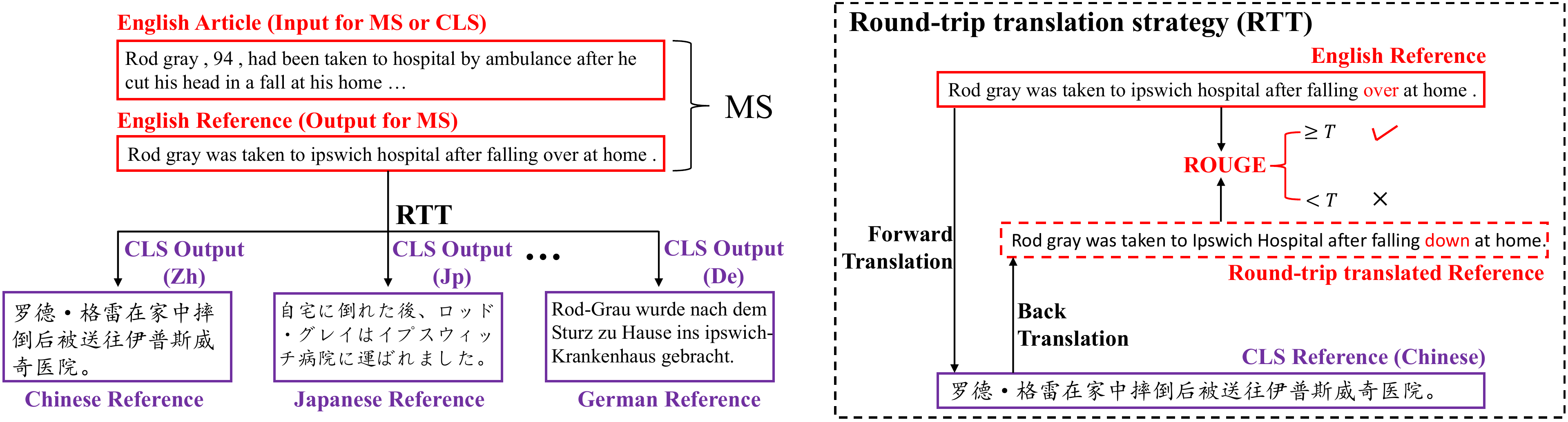}\\
  \caption{Overview of CLS corpora construction. Our method can be extended to many other language pairs and we focus on En2Zh and Zh2En in this paper. During RTT, we filter the sample in which ROUGE F1 score between the original reference and the round-trip translated reference is below a preset threshold $T$.}
  \label{data}
\end{figure*}

Our main contributions are as follows:
\begin{itemize}
	\item We propose a novel round-trip translation strategy to acquire large-scale CLS datasets from existing large-scale MS datasets. We have constructed a 370K English-to-Chinese (En2Zh) CLS corpus and a 1.69M Chinese-to-English (Zh2En) CLS corpus.
	\item To train the CLS systems in an end-to-end manner, we present neural cross-lingual summarization.  Furthermore, we propose to improve NCLS by incorporating MT and MS into CLS training process under multi-task learning. To the best of our knowledge, this is the first work to present an end-to-end CLS framework that trained on parallel corpora.
	\item Experimental results demonstrate that NCLS can achieve +4.87 ROUGE-2 on En2Zh and +5.07 ROUGE-2 on Zh2En over traditional pipeline paradigm. In addition, NCLS with multi-task learning can further achieve +3.60 ROUGE-2 on En2Zh and +0.72 ROUGE-2 on Zh2En. Our methods can be regarded as a benchmark for further NCLS studying.
\end{itemize}

\begin{table*}[t]
\centering
\setlength{\abovecaptionskip}{0.1cm}    
\setlength{\belowcaptionskip}{-0.3cm}
\begin{tabular}{@{}lrrrlrrr@{}}
\toprule
En2ZhSum         & \multicolumn{1}{c}{train} & \multicolumn{1}{c}{valid} & \multicolumn{1}{c}{test} & Zh2EnSum         & \multicolumn{1}{c}{train} & \multicolumn{1}{c}{valid} & \multicolumn{1}{c}{test} \\ \midrule
\#Documents      & 364,687                   & 3,000                     & 3,000                    & \#Documents      & 1,693,713                 & 3,000                     & 3,000                    \\
\#AvgWords (S)   & 755.09                    & 759.55                    & 744.84                   & \#AvgChars (S)   & 103.59                    & 103.56                    & 140.06                   \\
\#AvgEnWords (R) & 55.21                     & 55.28                     & 54.76                    & \#AvgZhChars (R) & 17.94                     & 18.00                     & 18.08                    \\
\#AvgZhChars (R) & 95.96                     & 96.05                     & 95.33                    & \#AvgEnWords (R) & 13.70                     & 13.74                     & 13.84                    \\
\#AvgSentsWords  & 19.62                     & 19.63                     & 19.61                    & \#AvgSentsChars  & 52.73                     & 52.41                     & 53.38                    \\
\#AvgSents       & 40.62                     & 41.08                     & 40.25                    & \#AvgSents       & 2.32                      & 2.33                      & 2.30                     \\ \bottomrule
\end{tabular}
\caption{Corpus statistics. \textbf{\#AvgWords (S)} is the average number of English words in the source document. Each reference has a bilingual version since each reference in CLS corpus is translated from the corresponding reference in the MS corpus. \textbf{\#AvgEnWords (R)} means the average number of words in English reference and \textbf{\#AvgZhChars (R)} denotes the average number of characters in Chinese reference. \textbf{\#AvgSentsWords} (\textbf{\#AvgSentsChars}) indicates the average number of words (characters) in a sentence in the source document. \textbf{\#AvgSents} refers to the average number of sentences in the source document.   \label{statistics}}
\end{table*}

\section{Dataset Construction} \label{dataconstruction}
Existing large-scale monolingual summarization datasets are automatically collected from the internet. CNN/Dailymail~\cite{hermann2015} dataset has been collected from \textit{CNN} and \textit{DailyMail} websites, where the article and news highlights are treated as the input and output respectively.  Similar to \citet{hermann2015}, \citet{Zhu2018MSMO} have constructed a multimodal summarization dataset MSMO where the text input and output are similar to that in CNN/Dailymail. We refer to the union set of CNN/Dailymail and MSMO as ENSUM\footnote{It contains 626,634 English summarization pairs.}. \citet{lcsts} introduce a large-scale corpus of Chinese short text summarization (LCSTS\footnote{It contains 2,400,591 Chinese summarization pairs.}) dataset constructed from the Chinese microblogging website \textit{Sina Weibo}. In this section, we introduce how to construct the En2Zh and Zh2En CLS datasets based on ENSUM and LCSTS respectively.


\textbf{Round-trip translation strategy.} Round-trip translation\footnote{\url{https://en.wikipedia.org/wiki/Round-trip_translation}} (RTT) is the process of translating a text into another language (forward translation), then translating the result back into the original language (back translation), using MT service\footnote{\url{http://www.anylangtech.com}}. Inspired by \citet{lample2018unsupervised}, we propose to adopt the round-trip translation to acquire CLS dataset from MS dataset. The process of constructing our corpora is shown in Figure~\ref{data}. 

Taking the construction of En2Zh corpus as an example, given a document-summary pair $(D_{\textrm{en}},S_{\textrm{en}})$, we first translate the summary $S_{\textrm{en}}$ into Chinese $S_{\textrm{zh}}$ and then back into English $S'_{\textrm{en}}$. 
The En2Zh document-summary pair $(D_{\textrm{en}},S_{\textrm{zh}})$, which satisfies $\textrm{ROUGE-1}(S_{\textrm{en}}, S'_{\textrm{en}}) \geqslant T_1$ and $\textrm{ROUGE-2}(S_{\textrm{en}}, S'_{\textrm{en}}) \geqslant T_2$ ($T_1$ is set to 0.45 for English and 0.6 for Chinese respectively, and $T_2$ is set to 0.2 here\footnote{The values are obtained by conducting a manual estimation on some samples randomly selected from two corpora.}), will be regarded as a positive pair. Otherwise, the pair will be filtered. Note that there are multiple sentences in $S_{\textrm{en}}$ in ENSUM, we apply the RTT to filter low-quality translated reference sentence by sentence. Once more than two-thirds of the sentences in the summary in a sample are retained, we will keep the sample. This process helps to ensure that the final compression ratio in our task does not differ too much from the actual compression ratio. Similar process is used on constructing Zh2En corpus. The ROUGE scores between Chinese sentences are calculated using Chinese characters as segmentation units.


\textbf{Corpus Statistics.} After conducting the round-trip translation strategy, we have obtained 370,759 En2Zh CLS pairs from ENSUM and 1,699,713 Zh2En CLS pairs from LCSTS. The statistics of En2Zh corpus (En2ZhSum) and Zh2En corpus (Zh2EnSum) are presented in Table~\ref{statistics}. In order to evaluate various CLS methods more reliably, we recruit 10 volunteers to correct the reference in the test sets in two constructed corpora.

\section{Approach}
The traditional approaches (Section~\ref{baseline}) intuitively treat CLS as a pipeline process which leads to error propagation. To handle that, we present the neural cross-lingual summarization methods (Section~\ref{single}), which train CLS in an end-to-end manner for the first time. Due to the strong relationship between CLS, MS, and MT tasks, we propose to incorporate MS and MT into CLS training under multi-task learning (Section~\ref{multi}).

\subsection{Baseline Pipeline Methods} \label{baseline}
In general, traditional CLS is composed of summarization step and translation step. The different order of these two steps leads to the following two strategies. Take En2Zh CLS as an example. 

\textbf{Early Translation (ETran).} This strategy first translates the English document to Chinese document with machine translation. Then a Chinese summary is generated by a summarization model.

\textbf{Late Translation (LTran).} This strategy first summarizes the English document to a short English summary and then translates it into Chinese.

\begin{figure}[t]
\setlength{\abovecaptionskip}{0.11cm}    
\setlength{\belowcaptionskip}{-0.4cm}
  \centering
  \includegraphics[width=6cm]{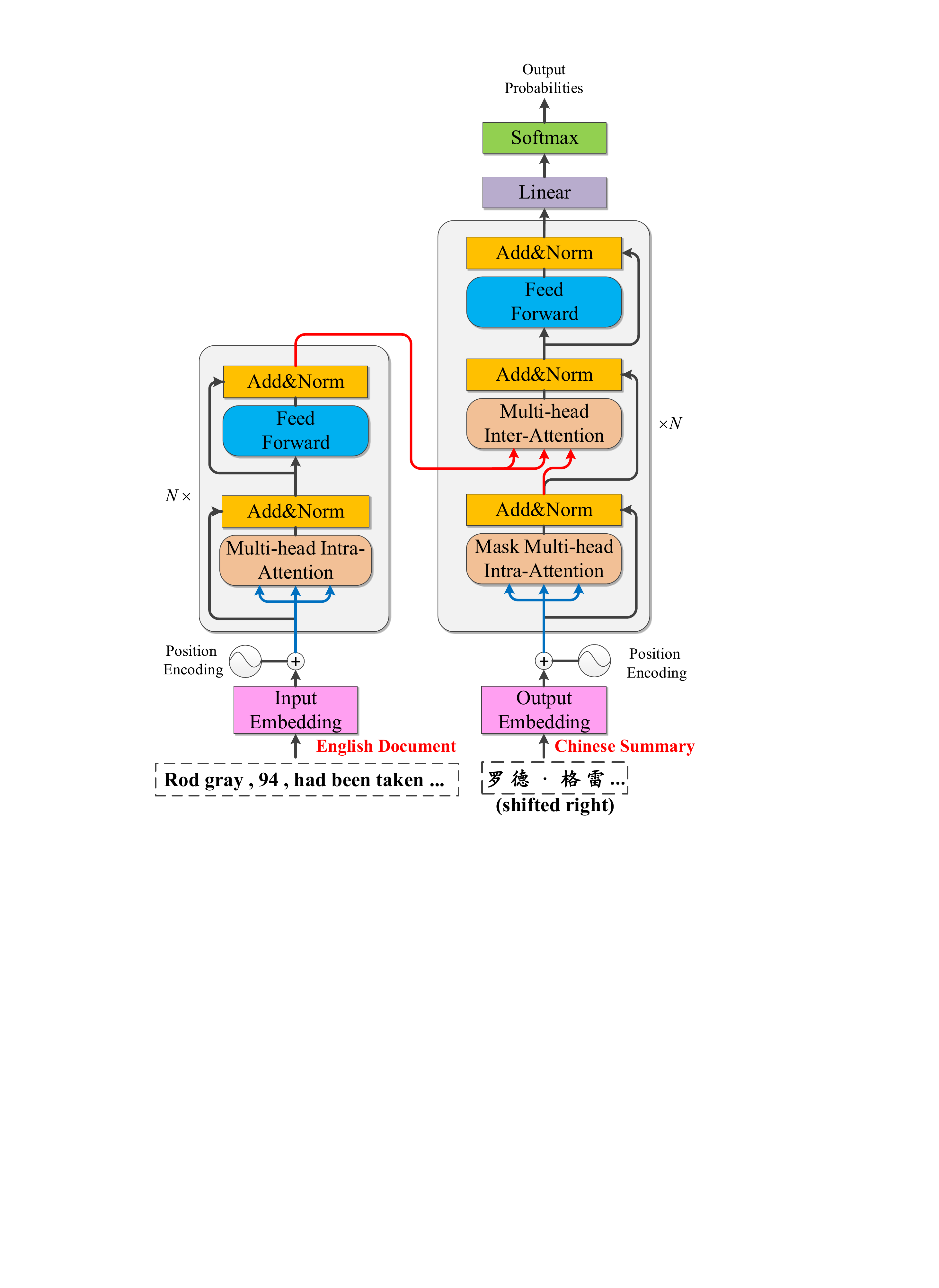}\\
  \caption{Transformer-based NCLS models (TNCLS).}
  \label{trans-ncls}
\end{figure}

\subsection{Neural Cross-Lingual Summarization} \label{single}
Considering the excellent text generation performance of Transformer encoder-decoder network~\cite{transformer}, we implement our NCLS models entirely based on this framework in this work. As shown in Figure~\ref{trans-ncls}, given a set of CLS data $D={(X^{(i)},Y^{(i)})}$ where both $X$ and $Y$ are a sequence of tokens, the encoder maps the input document $X=(x_1,x_2,\cdots,x_n)$ into a sequence of continuous representations $z=(z_1,z_2,\cdots,z_n)$ whose size varies with respect to the source sequence length. The decoder generates a summary $Y=(y_1,y_2,\cdots,y_m)$, which is in a different language, from the continuous representations. The encoder and decoder are trained jointly to maximize the conditional probability of target sequence given a source sequence:
\begin{equation} 
\small
\setlength\abovedisplayskip{0.1cm}
\setlength\belowdisplayskip{0.1cm}
	L_{\theta} = \sum_{t=1}^{N}\rm{log}P(y_t|y_{<t},x;\theta)
\end{equation}

Transformer is composed of stacked encoder and decoder layers. Consisting of two blocks, the encoder layer is a self-attention block followed by a position-wise feed-forward block. Despite the same architecture as the encoder layer, the decoder layer has an extra encoder-decoder attention block. Residual connection and layer normalization are used around each block. In addition, the self-attention block in the decoder is modified with masking to prevent present positions from attending to future positions during training.

For self-attention and encoder-decoder attention, a multi-head attention block is used to obtain information from different representation subspaces at different positions. Each head corresponds to a scaled dot-product attention, which operates on the query $Q$, key $K$, and value $V$:
\begin{equation}
\small
\setlength\abovedisplayskip{0.15cm}
\setlength\belowdisplayskip{0.15cm}
	\textrm{Attention}(Q,K,V) = \textrm{softmax}(\frac{QK^T}{\sqrt{d_k}})V
\end{equation}
where $d_k$ is the dimension of the key.

Finally, the output values are concatenated and projected by a feed-forward layer to get final values:
\begin{equation}
\small
\setlength\abovedisplayskip{0.15cm}
\setlength\belowdisplayskip{0.15cm}
\begin{split}
\textrm{MultiHead}(Q,K,V) &= \textrm{Concat}(\textrm{head}_1,\ldots,\textrm{head}_h) W^O  \\
\textrm{where}\ \textrm{head}_i &= \textrm{Attention}(QW_i^Q, KW_i^K, VW_i^V)
\end{split}
\end{equation}
where $W^O$, $QW_i^Q$, $KW_i^K$, and $VW_i^V$ are learnable matrices, $h$ is the number of heads.

\subsection{Improving NCLS with MS and MT} \label{multi} 
Considering there is a strong relationship between CLS task and MS task, as well as between CLS task and MT task: (1) CLS shares the same goal with MS, i.e., to grasp the core idea of the original document, but the final results are presented in different languages. (2) From the perspective of information compression, machine translation can be regarded as a special kind of cross-lingual summarization with a compression ratio of 1:1. Therefore, we consider using MS and MT datasets to further improve the performance of CLS task under multi-task learning. 

\begin{figure}[t]
\setlength{\abovecaptionskip}{0.11cm}    
\setlength{\belowcaptionskip}{-0.4cm}
  \centering
  \includegraphics[width=7.5cm]{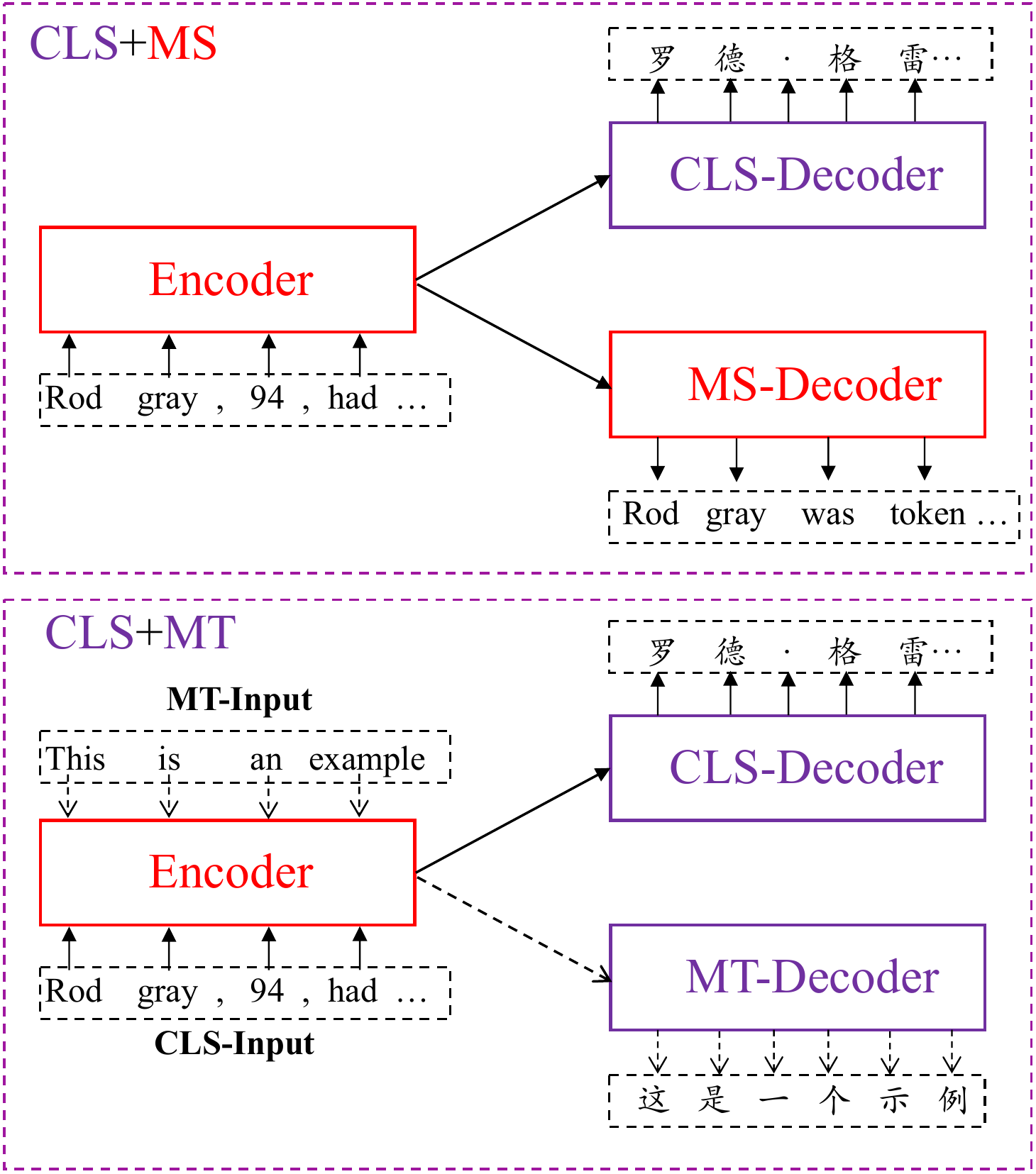}\\
  \caption{Overview of multi-task NCLS. The lower half is CLS+MT using alternating training strategy. Different colors represent different languages.}
  \label{mncls}
\end{figure}

Inspired by \citet{luong2016}, we employ the one-to-many scheme to incorporate the training process of MS and MT into that of CLS. As shown in Figure~\ref{mncls}, this scheme involves one encoder and multiple decoders for tasks in which the encoder can be shared. We study two different task combinations here: CLS+MS and CLS+MT.

\textbf{CLS+MS.} Note that the reference in each of CLS datasets has a bilingual version. For instance, En2ZhSum dataset contains a total of 370,687 documents with corresponding summaries in both Chinese and English. Thus, we consider jointly training CLS and MS as follows. Given a source document, the encoder encodes it into continuous representations, and then the two decoders simultaneously generate the output of their respective tasks. The loss can be calculated as follows:
\begin{equation}
\small
\setlength\abovedisplayskip{0.15cm}
\setlength\belowdisplayskip{0.15cm}
	L_{\theta} = \sum_{t=1}^{N^{(1)}}\rm{log}P(y^{(1)}_t|y^{(1)}_{<t},x;\theta) + \sum_{t=1}^{N^{(2)}}\rm{log}P(y^{(2)}_t|y^{(2)}_{<t},x;\theta)
\end{equation}
where $y^{(1)}$ and $y^{(2)}$ are the outputs of two tasks.


\textbf{CLS+MT.} Since CLS input-output pairs are different from MT input-output pairs, we consider adopting the alternating training strategy~\cite{dong2015multi}, which optimizes each task for a fixed number of mini-batches before switching to the next task, to jointly train CLS and MT. For MT task, we employ 2.08M\footnote{LDC2000T50, LDC2002L27, LDC2002T01, LDC2002E18, LDC2003E07, LDC2003E14, LDC2003T17, LDC2004T07} sentence pairs from LDC corpora with CLS dataset to train CLS+MT.

\section{Experiments}
\subsection{Experimental Settings}
 For English, we apply two different granularities of segmentation, i.e., words and subwords~\cite{bpe}. We lowercase all English characters. We truncate the input to 200 words and the output to 120 words (150 characters for Chinese output) . For Chinese, we employ three different granularities of segmentation: characters, words, and subwords. It is worth noting that we only apply subword-based segmentation in Zh2En model since subword-based segmentation will make the English article much longer in En2Zh (especially at the Chinese target-side output), which makes the Transformer performs extremely poor. For our baseline pipeline models, the vocabulary size of Chinese characters is 10,000, and that of Chinese words, Chinese subwords, and English words are all 100,000. In our En2Zh NCLS models, the vocabulary size of source-side English words is 100,000, and that of target-side Chinese characters and words are 18,000, and 50,000 respectively. In our Zh2En models, the vocabulary size of source-side Chinese characters, words, and subwords are 10,000, 100,000, and 100,000 respectively, and that of target-side English words and subwords are all 40,000. We initialize all the parameters via Xavier initialization methods~\cite{Glorot10understandingthe}. We train our models using configuration \textit{transformer\_base}~\cite{transformer}, which contains a 6-layer encoder and a 6-layer decoder with 512-dimensional hidden representations. 
 
 During training, in En2Zh models, each mini-batch contains a set of document-summary pairs with roughly 2,048 source and 2,048 target tokens; in Zh2En models, each mini-batch contains a set of document-summary pairs with roughly 4,096 source and 4,096 target tokens. We use Adam optimizer~\cite{kingma2014adam} with $\beta_1=0.9$, $\beta_2=0.998$, and $\epsilon=10^{-9}$. We use a single NVIDIA TITAN X to train our models. Convergence is reached within 1,000,000 iterations in both TNCLS models and baseline models. We train each task for about 800,000 iterations in multi-task NCLS models (reaching convergence). At test time, our summaries are produced using beam search with beam size 4. 

\begin{table}[t]
\small
\setlength{\abovecaptionskip}{0.1cm}    
\setlength{\belowcaptionskip}{-1cm}
\begin{tabular}{@{}lccc@{}}
\toprule
Model & ROUGE-1 & ROUGE-2 & ROUGE-L \\ \midrule
\citet{gu2016copy} & 35.00 & 22.30 & 32.00 \\
\citet{piji2017-deep} & 36.99 & 24.15 & 34.21 \\
Transformer & 39.71 & 27.45 & 37.13 \\ \bottomrule
\end{tabular}
\caption{Performance of our implemented transformer-based monolingual summarization model on LCSTS. \label{base1}}
\end{table}

\subsection{Baselines and Model Variants}
We compare our NCLS models with the following two traditional methods:

\textbf{TETran:} We first translate the source document via a Transformer-based machine translation model trained on LDC corpora. Then we employ LexRank~\cite{erkan2004lexrank}, a strong and widely used unsupervised summarization method, to summarize the translated document. The reason why we choose to apply an unsupervised method is that we lack the version of MS dataset in the target language to train a supervised model to summarize the translated document.

\textbf{TLTran:} We first build a Transformer-based MS model which is trained on the original MS dataset. Then the MS model aims to summarize the source document into a summary. Finally, we translate the summary into target language by using the Transformer-based machine translation model trained on LDC corpora. The performance of our transformer-based MS models is given in Table~\ref{base1} and Table~\ref{base2}.

To make our experiments more comprehensive, during the process of TETran and TLTran, we replace the Transformer-based machine translation model with Google Translator\footnote{\url{https://translate.google.com/}}, which is one of the state-of-the-art machine translation systems. We refer to these two methods as \textbf{GETran} and \textbf{GLTran} respectively.

There are three variants of our NCLS models:

\textbf{TNCLS:} Transformer-based NCLS models where the input and output are different granularities combinations of units.

\textbf{CLS+MS:} It refers to the multi-task NCLS model which accepts an input text and simultaneously performs text generation for both CLS and MS tasks and calculates the total losses.

%

\textbf{CLS+MT:} It trains CLS and MT tasks via alternating training strategy. Specifically, we optimize the CLS task in a mini-batch, and we optimize the MT task in the next mini-batch.


\begin{table}[t]
\small
\setlength{\abovecaptionskip}{0.1cm}    
\setlength{\belowcaptionskip}{-0.6cm}
\begin{tabular}{@{}lccc@{}}
\toprule
Model & ROUGE-1 & ROUGE-2 & ROUGE-L \\ \midrule
\citet{see2017} & 39.53 & 17.28 & 36.38 \\
Transformer & 39.24 & 16.67 & 36.42 \\ \bottomrule
\end{tabular}
\caption{Performance of our implemented transformer-based MS model on CNN/Dailymail. \label{base2}}
\end{table}

\subsection{Experimental Results and Analysis}

\begin{table*}[]
\centering
\setlength{\abovecaptionskip}{0.1cm}    
\setlength{\belowcaptionskip}{-0.1cm}
\begin{tabular}{@{}lccccc@{}}
\toprule
\multirow{2}{*}{Model}  & \multirow{2}{*}{Unit} & En2ZhSum                   & En2ZhSum*                  & Zh2EnSum                   & Zh2EnSum*                  \\ \cmidrule(l){3-6} 
                        &                       & RG1-RG2-RGL($\uparrow$)    & RG1-RG2-RGL($\uparrow$)    & RG1-RG2-RGL($\uparrow$)    & RG1-RG2-RGL($\uparrow$)    \\ \midrule
TETran                  & --                    & \textbf{26.12-10.59-23.21} & \textbf{26.15-10.60-23.24} & \textbf{22.81-\, 7.17-18.55} & \textbf{23.09-\, 7.33-18.74} \\ \midrule
GETran                  & --                    & \textbf{28.17-11.38-25.75} & \textbf{28.19-11.40-25.77} & \textbf{24.03-\, 8.91-19.92} & \textbf{24.34-\, 9.14-20.13} \\ \midrule
\multirow{3}{*}{TLTran} & c-c                   & --                         & --                         & 32.85-15.34-29.21          & 33.01-15.43-29.32          \\
                        & w-w                   & \textbf{30.20-12.20-27.02} & \textbf{30.22-12.20-27.04} & 31.11-13.23-27.55          & 31.38-13.42-27.69          \\
                        & sw-sw                 & --                         & --                         & \textbf{33.64-15.58-29.74} & \textbf{33.92-15.81-29.86} \\ \midrule
\multirow{3}{*}{GLTran} & c-c                   & --                         & --                         & 34.44-15.71-30.13          & 34.58-16.01-30.25          \\
                        & w-w                   & \textbf{32.15-13.84-29.42} & \textbf{32.17-13.85-29.43} & 32.42-15.19-28.75          & 32.52-15.39-28.88          \\
                        & sw-sw                 & --                         & --                         & \textbf{35.28-16.59-31.08} & \textbf{35.45-16.86-31.28} \\ \midrule
\multirow{4}{*}{TNCLS}  & c-w                   & --                         & --                         & 36.36-19.74-32.66          & 35.82-19.04-32.06          \\
                        & w-c                   & \textbf{36.83-18.76-33.22} & \textbf{36.82-18.72-33.20} & --                         & --                         \\
                        & w-w                   & 33.09-14.85-29.82          & 33.10-14.83-29.82          & 38.54-22.34-35.05          & 37.70-21.15-34.05          \\
                        & sw-sw                 & --                         & --                         & \textbf{39.80-23.15-36.11} & \textbf{38.85-21.93-35.05} \\ \bottomrule
\end{tabular}
\caption{ROUGE F1 scores (\%) on En2ZhSum and Zh2EnSum test sets. En2ZhSum* and Zh2EnSum* are the corresponding human-corrected test sets. \textit{Unit} denotes the granularity combination of text units, where \textit{c} means character, \textit{w} means word, and \textit{sw} means subword. RG refers to ROUGE for short. $\uparrow$ indicates that the larger values, the better the results are. Our NCLS models perform significantly better than baseline models by the 95\% confidence interval measured by the official ROUGE script\protect\footnotemark. \label{result1}}
\end{table*}
\footnotetext{The parameter for ROUGE script here is ``-c 95 -r 1000 -n 2 -a".}

\textbf{Comparison between NCLS with baselines.} We evaluate different models with the standard ROUGE metric~\cite{lin2004rouge}, reporting the F1 scores for ROUGE-1, ROUGE-2, and ROUGE-L. The results are presented in Table~\ref{result1}.

We can find that GLTran outperforms TLTran and GETran outperforms TETran, which indicates that pipeline-based methods perform better when using a stronger machine translation system. Compared with GLTran or GETran, our TNCLS models both achieve significant improvements, which can verify our motivation and demonstrate the efficacy of our constructed corpora. 

\begin{table*}[t]
\small
\setlength{\abovecaptionskip}{0.1cm}    
\centering
\begin{tabular}{@{}lccccc@{}}
\toprule
\multirow{2}{*}{DataVersion} & \multicolumn{1}{l}{\multirow{2}{*}{BT?}} & En2ZhSum & En2ZhSum* & Zh2EnSum & Zh2EnSum* \\ \cmidrule(l){3-6} 
 & \multicolumn{1}{l}{} & RG1-RG2-RGL($\uparrow$) & RG1-RG2-RGL($\uparrow$) & RG1-RG2-RGL($\uparrow$) & RG1-RG2-RGL($\uparrow$) \\ \midrule
Filter & YES & 36.83-18.76-33.22 & 36.82-18.72-33.20 & 39.80-23.15-36.11 & 38.85-21.93-35.05 \\
Pseudo-Filter & NO & 36.04-17.80-32.49 & 36.03-17.78-32.48 & 35.58-17.93-31.71 & 35.00-17.37-31.10 \\ \midrule
Non-Filter & NO & 37.62-19.88-33.99 & 37.62-19.85-33.99 & 36.51-19.23-32.77 & 36.03-18.63-32.19 \\ \bottomrule
\end{tabular}
\caption{Experimental results on different versions of datasets. \textit{Filter} refers to the version of dataset for which we employ RTT strategy to filter. \textit{Non-Filter} denotes the version of the dataset obtained by simply forward translation without filtering process including back translation. \textit{Pseudo-Filter} is the dataset randomly sampled from \textit{Non-Filter} version and is of the same size as \textit{Filter} version. BT refers to back translation in RTT. For En2Zh task, we train the TNCLS (\textit{w}-\textit{c}). For Zh2En task, we train the TNCLS (\textit{sw}-\textit{sw}). 
\label{result2}}
\end{table*}

In En2Zh CLS task, the results of each model on En2ZhSum are similar to those on En2ZhSum*. This is because the original ENSUM dataset comes from the news reports. Existing MT for news reports has excellent performance. Besides, we have pre-filtered samples with low translation quality during dataset construction. Therefore, the quality of the automatic test set is high. TNCLS (\textit{w}-\textit{c}) performs significantly better than TNCLS (\textit{w}-\textit{w}). This is because the character-based segmentation can greatly reduce the vocabulary size at the Chinese target-side, which leads to generating nearly no UNK token during the decoding process.

In Zh2En CLS task, the subword-based models outperform others since subword-based segmentation can greatly reduce the vocabulary size and the generation of UNK. Compared with baselines, TNCLS can achieve maximum improvement up to \textbf{+4.52 ROUGE-1}, \textbf{+6.56 ROUGE-2}, \textbf{+5.03 ROUGE-L} on Zh2EnSum and \textbf{+3.40 ROUGE-1}, \textbf{+5.07 ROUGE-2}, \textbf{+3.77 ROUGE-L} on Zh2EnSum*. The results of TNCLS drops obviously on the human-corrected test set, showing that the quality of the translated reference is not as perfect as expected. The reason is straightforward that the original LCSTS dataset comes from social media so that the proportion of abbreviations and omitting punctuation in its text is much higher than in news, resulting in lower translation quality. 

In conclusion, TNCLS models significantly outperform the traditional pipeline methods on both En2Zh and Zh2En CLS tasks.

\begin{table*}[h]
\small
\setlength{\abovecaptionskip}{0.1cm}    
\centering
\begin{tabular}{@{}lcccc@{}}
\toprule
\multirow{2}{*}{Model} & En2ZhSum & En2ZhSum* & Zh2EnSum & Zh2EnSum* \\ \cmidrule(l){2-5} 
 & RG1-RG2-RGL($\uparrow$) & RG1-RG2-RGL($\uparrow$) & RG1-RG2-RGL($\uparrow$) & RG1-RG2-RGL($\uparrow$) \\ \midrule
TNCLS & \textbf{36.83-18.76-33.22} & \textbf{36.82-18.72-33.20} & \textbf{39.80-23.15-36.11} & \textbf{38.85-21.93-35.05} \\ \midrule
CLS+MS & 38.23-20.21-34.76 & 38.25-20.20-34.76 & 41.08-23.67-37.19 & \textbf{40.34-22.65-36.39} \\
CLS+MT & \textbf{40.24-22.36-36.61} & \textbf{40.23-22.32-36.59} & \textbf{41.09-23.70-37.17} & 40.25-22.58-36.21 \\ \bottomrule
\end{tabular}
\caption{Results of multi-task NCLS. The granularity combination of input and output in En2Zh task is ``word to character" (\textit{w}-\textit{c}), and that in Zh2En task is ``subword to subword" (\textit{sw}-\textit{sw}).
\label{result3}}
\end{table*}

\textbf{Why Back Translation?} To show the influence of filtering the corpus by back translation during the RTT process, we use three kinds of datasets to train our TNCLS models and compare their performance. They are: (a) the CLS dataset obtained by simply employing forward translation on MS dataset (\textit{Non-Filter}); (b) the CLS dataset obtained by a complete RTT process (\textit{Filter}); (c) the dataset obtained by sampled from \textit{Non-Filter} dataset to keep the same size as the \textit{Filter} dataset (\textit{Pseudo-Filter}). The results are given in Table~\ref{result2}. The models trained on \textit{Filter} dataset significantly outperform the models trained on \textit{Pseudo-Filter} dataset on both En2Zh and Zh2En tasks, which indicates that the back translation can effectively filter dirty samples and improve the overall quality of corpora, thus boosting the performance of NCLS. In En2Zh task, the model trained on \textit{Non-Filter} dataset performs best. The reasons are two-fold: (1) the quality of machine translation for English news is reliable; (2) the scale of \textit{Non-Filter} dataset is almost twice that of the two others so that after the amount of data reaches a certain level, it can make up for the noises caused by the translation error in the corpus. In Zh2En task, the performance of the model trained on \textit{Non-Filter} dataset is not as good as that on \textit{Filter}. It can be attributed to the fact that current MT is not very ideal in the translation of texts on social media so that the dataset constructed by only using forward translation contains too many noises. Therefore, when the quality of machine translation is not that ideal, backward translation is especially important during the process of constructing corpus.

\textbf{Results of Multi-task NCLS.} To explore whether MS and MT can further improve NCLS, we compare the multi-task NCLS with NCLS using one same granularity combination of units. The results are given in Table~\ref{result3}. As shown in Table~\ref{result3}, both CLS+MS and CLS+MT can improve the performance of NCLS, which can be attributed to that the encoder is enhanced by incorporating MS and MT data into the training process. CLS+MT significantly outperforms CLS+MS in En2Zh task while CLS+MS performs comparably with CLS+MT in Zh2En task. The reasons are two-fold: (1) In En2Zh task, MT dataset is much larger than both MS and CLS datasets, which makes it more necessary for enhancing the robustness of encoder. (2) We use the LDC MT dataset, which belongs to the news domain similar to our En2ZhSum, during the training of CLS+MT. However, Zh2EnSum belongs to social media domain, thus resulting in the greater improvement of CLS+MT in En2Zh than in Zh2En. In general, NCLS with multi-task learning achieves more significant improvement in En2Zh task than in Zh2En task, which illustrates that extra dataset in other related tasks is essentially important for boosting the performance when CLS dataset is not very large.



\textbf{Human Evaluation.} We conduct the human evaluation on 25 random samples from each of the En2ZhSum and Zh2EnSum test set. We compare the summaries generated by our methods (including TNCLS, CLS+MS, and CLS+MT) with the summaries generated by GLTran. Three graduate students are asked to compare the generated summaries with human-corrected references, and assess each summary from three independent perspectives: (1) How informative the summary is? (2) How concise the summary is? (3) How fluent, grammatical the summary is? Each property is assessed with a score from 1 (worst) to 5 (best). The average results are presented in Table~\ref{human}.

\begin{table}[]
\small
\setlength{\abovecaptionskip}{0.1cm}    
\setlength{\belowcaptionskip}{-0.5cm}
\centering
\begin{tabular}{@{}lcccccc@{}}
\toprule
\multirow{2}{*}{Model} & \multicolumn{3}{c}{En2Zh} & \multicolumn{3}{c}{Zh2En} \\ \cmidrule(l){2-7} 
 & IF & CC & FL & IF & CC & FL \\ \midrule
GLTran & 3.06 & 3.37 & 3.13 & 3.53 & 4.21 & 4.25 \\
TNCLS & 3.25 & 3.33 & 3.17 & 3.67 & 4.25 & 4.24 \\
CLS+MS & 3.53 & 3.58 & 3.53 & 3.72 & 4.31 & 4.28 \\
CLS+MT & \textbf{3.58} & \textbf{3.76} & \textbf{3.63} & \textbf{3.78} & \textbf{4.43} & \textbf{4.35} \\ \bottomrule
\end{tabular}
\caption{Human evaluation results. IF, CC and FL denote informative, concise, and fluent respectively.}
\label{human}
\end{table}

\begin{figure*}[t]
\setlength{\abovecaptionskip}{0.11cm}    
\setlength{\belowcaptionskip}{-0.5cm}
  \centering
  \includegraphics[width=1\linewidth]{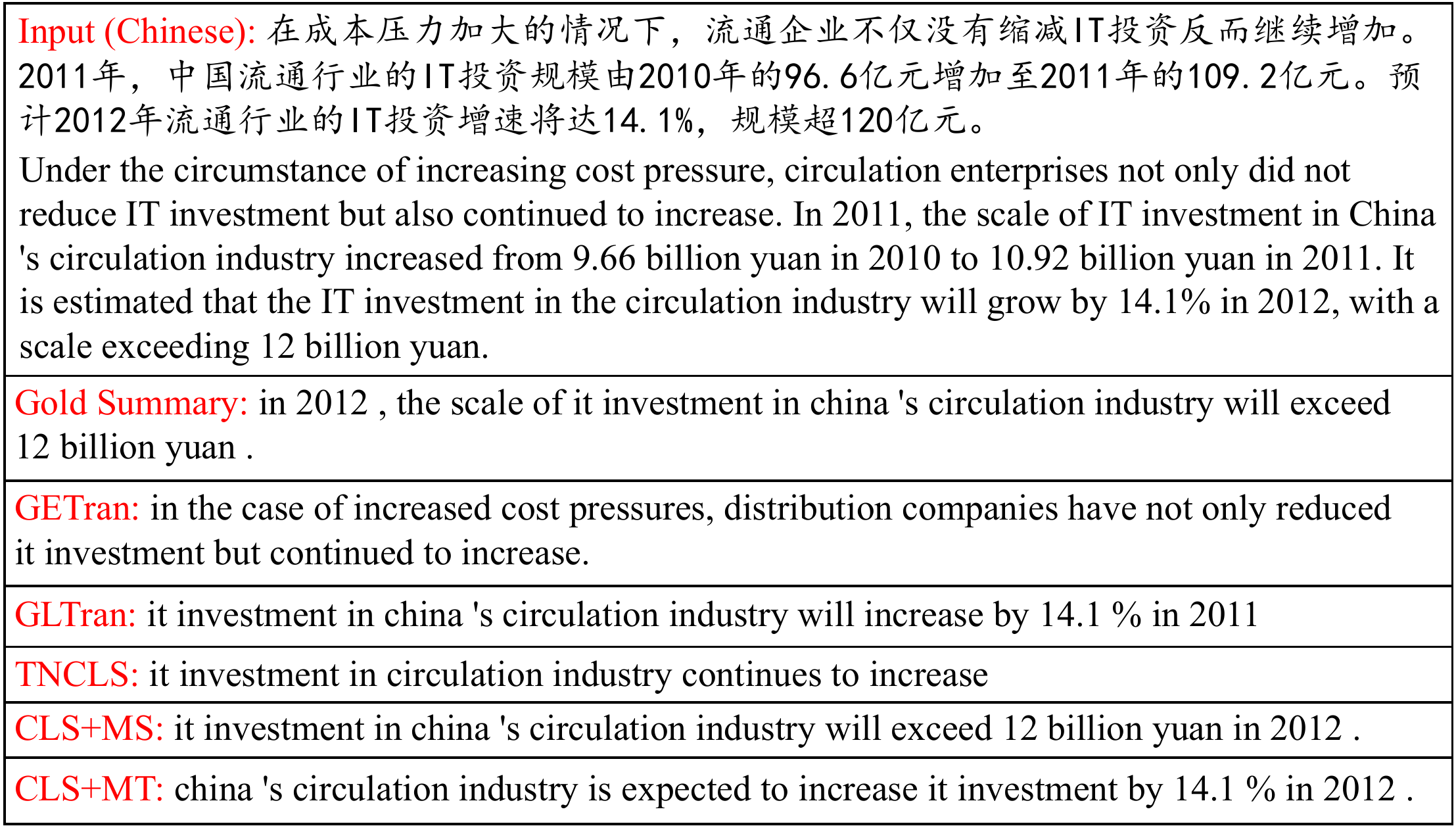}\\
  \caption{Examples of generated summaries.}
  \label{example}
\end{figure*}

As shown in Table~\ref{human}, TNCLS can generate more informative summaries compared with GLTran, which shows the advantage of end-to-end models. The conciseness score and fluency score of TNCLS are comparable to those of GLTran. This is because both GLTtrans and TNCLS employ a single encoder-decoder model, which easily leads to under-generation and repetition. Our CLS+MS and CLS+MT can significantly improve the conciseness and fluency of generated summaries, which shows that these methods can generate shorter summaries and reduce grammatical errors. In conclusion, TNCLS can generate more informative summaries, but it is difficult to improve the conciseness and fluency. However, with the help of MT and MS tasks, conciseness and fluency scores can be significantly improved.

\subsection{Case Study}
We show the case study of a sample from the Zh2EnSum human-corrected test set in Figure~\ref{example}. As shown in Figure~\ref{example}, the summary generated by \textbf{GETran} obviously suffers from errors of machine translation (``distribution companies" should be corrected as ``circulation enterprises"). Since \textbf{GETran} first translates all the source text, it is easier to bring the errors from machine translation. The \textbf{GLTran}-generated summary contracts the fact that the year in it should be 2012 instead of 2011. The translation quality of the sentence is relatively reliable, thus the errors are probably produced during the summarization step. Compared with the first two generated summaries, although the summary produced by \textbf{TNCLS} does not emphasize the time and place of occurrence, there is no mistake in the logic of its expression. The summaries generated by \textbf{CLS+MS} and \textbf{CLS+MT} are generally consistent with the facts, but their emphases are different. The \textbf{CLS+MS} summary matches the gold summary better. The flaws of both of them are that they do not reflect the ``scale" in the original text. In conclusion, our methods can produce more accurate summaries than baselines.

\section{Related Work}
Cross-lingual summarization has been proposed to present the most salient information of a source document in a different language, which is very important in the field of multilingual information processing. Most of the existing methods handle the task of CLS via simply applying two typical translation schemes, i.e., early translation~\cite{leuski2003cross, ouyang2019robust} and late translation~\cite{orasan2008evaluation, wan2010cross}. The early translation scheme first translates the original document into target language and then generates the summary of the translated document. The late translation scheme first summarizes the original document into a summary in the source language and then translates it into target language.

\citet{leuski2003cross} translate the Hindi document to English and then generate the English headline for it. \citet{ouyang2019robust} present a robust abstractive summarization system for low resource languages where no summarization corpora are currently available. They train a neural abstractive summarization model on noisy English documents and clean English reference summaries. Then the model can learn to produce fluent summaries from disfluent inputs, which allows generating summaries for translated documents.
\citet{orasan2008evaluation} summarize the Romanian news with the maximal marginal relevance method~\cite{goldstein2000multi} and produce the English summaries for English speakers. \citet{wan2010cross} adopt the late translation scheme for the task of English-to-Chinese CLS. They extract English sentences considering both the informativeness and translation quality of sentences and automatically translate the English summary into the final Chinese summary. The above researches only make use of the information from only one language side. 

Some methods have been proposed to improve CLS with bilingual information. \citet{wan2011using} proposes two graph-based summarization methods to leverage both the English-side and Chinese-side information in the task of English-to-Chinese CLS. Inspired by the phrase-based translation models, \citet{yao2015phrase} introduce a compressive CLS, which simultaneously performs sentence selection and compression. They calculate the sentence scores based on the aligned bilingual phrases obtained by MT service and perform compression via deleting redundant or poorly translated phrases. \citet{zhang2016abstractive} propose an abstractive CLS which constructs a pool of bilingual concepts represented by the bilingual elements of the source-side predicate-argument structures (PAS) and the target-side counterparts. The final summary is generated by maximizing both the salience and translation quality of the PAS elements.

However, all these researches belong to the pipeline paradigm which not only relies heavily on hand-crafted features but also causes error propagation. End-to-end deep learning has proven to be able to alleviate these two problems, while it has been absent due to the lack of large-scale training data. 
Recently, \citet{liu2018zero} present zero-shot cross-lingual headline generation based on existing parallel corpora of translation and monolingual headline generation. Similarly, \citet{duan-etal-2019-zero} propose to use monolingual abstractive sentence summarization system to teach zero-shot cross-lingual abstractive sentence summarization on both summary word generation and attention. Although great efforts have been made in cross-lingual summarization, how to automatically build a high-quality large-scale cross-lingual summarization dataset remains unexplored.

In this paper, we focus on English-to-Chinese and Chinese-to-English CLS and try to automatically construct two large-scale corpora respectively. In addition, based on the two corpora, we perform several end-to-end training methods noted as Neural Cross-Lingual Summarization.

\section{Conclusion and Future Work}
In this paper, we present neural cross-lingual summarization for the first time. To achieve that goal, we propose to acquire large-scale supervised data from existing monolingual summarization datasets via round-trip translation strategy. Then we apply end-to-end methods on our constructed datasets and find our NCLS models significantly outperform the traditional pipeline paradigm. Furthermore, we consider utilizing machine translation and monolingual summarization to further improve NCLS. Experimental results have shown that both machine translation and monolingual summarization can significantly help NCLS generate better summaries. 

In our future work, we will adopt our RTT strategy to obtain CLS datasets of other language pairs, such as English-to-Japanese, English-to-German, Chinese-to-Japanese, and Chinese-to-German, etc.

\section{Acknowledgments}
The research work described in this paper has been supported by the National Key Research and Development Program of China under Grant No. 2016QY02D0303. We thank the three anonymous reviewers for their careful reading of our paper and their many insightful comments and suggestions. We would like to thank He Bai, Yuchen Liu, Haitao Lin, Yang Zhao, Cong Ma, Lu Xiang, Weikang Wang, Zhen Wang, and Jiaqi Liang for their invaluable contributions in shaping the early stage of this work. We thank Xina Fu, Jinliang Lu, and Sikai Liu for conducting human evaluation. 


\bibliography{emnlp-ijcnlp-2019}
\bibliographystyle{acl_natbib}

\end{document}